\documentclass[pdflatex,sn-mathphys]{sn-jnl}
\jyear{2021}%

\theoremstyle{thmstyleone}%
\theoremstyle{thmstyletwo}%

\theoremstyle{thmstylethree}%

\begin{document}

\title[PSO for 3D medical image registration]{Particle Swarm Optimization in 3D Medical Image Registration: A Systematic Review}


\author*{\fnm{Lucia} \sur{Ballerini}}\email{lucia.ballerini@unistrapg.it}



\affil{
\orgname{University for Foreigners of Perugia}, 
\orgaddress{
\city{Perugia}, \postcode{06123}, 
\country{Italy}}}

\abstract{Medical image registration seeks to find an optimal spatial transformation that best aligns the underlying anatomical structures. These problems usually require the optimization of a similarity metric. Swarm Intelligence techniques are very effective and efficient optimization methods. This systematic review focuses on 3D medical image registration using Particle Swarm Optimization.}


\maketitle

\section{Introduction}\label{sec1}

In medical imaging, you may need to compare scans of multiple patients or scans of the same patient, using mono-modal or multi-modal images taken in different sessions~\cite{Hill_2001}. Moreover, images taken from different modalities provides complementary information.
To accomplish this, we need to align multiple medical images, volumes, or surfaces to a common coordinate system.
This process is called Image Registration.  

Huge number of algorithms are available for image registration. Indeed several review papers on this topic have been published. A simple search on PubMed for Review articles using the terms "Image Registration" gives over 1000 results. 
The field of medical image registration has been evolving rapidly with hundreds of papers published each year.
Though medical image registration has been extensively studied, it remains a hot research topic. 

Medical image registration seeks to find an optimal spatial transformation that best aligns the underlying anatomical structures. It usually requires the optimization of some similarity metric between the images. Among the optimization methods, the techniques based on Swarm Intelligence represent an efficient solution. 

The aim of this paper is to provide a comprehensive systematic review on medical image registration using Particle Swarm Optimization (PSO), which is a population-based stochastic optimization algorithm~\cite{Kennedy1995}.
This study concentrates on 3D medical images.
Though a few review papers on medical image registration using PSO have been published~\cite{Rundo2017,Saiti2020}, to the best of our knowledge this is the first systematic review on the subject.

\section{Medical Image Registration}\label{reg}

Medical image registration is the process of aligning or registering two or more medical images of the same or different modalities so that they can be compared and analyzed. This is often done to compare images taken at different times or from different viewpoints, or to combine information from multiple images to create a more detailed or accurate image.

There are several approaches to medical image registration, including intensity-based methods, feature-based methods, and deformable registration. Intensity-based methods use the intensity values of the pixels in the images to register the images, while feature-based methods use distinctive features in the images, such as edges or landmarks, to register the images. Deformable registration involves warping one image to match the other, and can be used when there are large differences between the images that cannot be compensated for by simple translations or rotations.

Medical image registration has a wide range of applications, including image-guided surgery, treatment planning, disease diagnosis and monitoring, and research. It is an important tool in medical imaging and has the potential to improve patient care and outcomes.

There have been many research papers published on the topic of medical image registration. Several surveys have been published.
For an extensive review of medical image registration techniques, we refer to Maintz and Viergever~\cite{Maintz1998} and Zitová and Flusser~\cite{Zitova2003}.





Intensity-based image registration methods use the intensity values of the pixels in the images to register the images. One common approach is to use Mutual Information (MI), which is a measure of the statistical dependence between the intensity values in the two images. Normalized Mutual Information (NMI) improves the robustness of MI by avoiding some mis-registrations by being independent of overlapping areas of the two datasets. 

Feature-based Registration methods aim to find the transformation that minimizes the distance between the features extracted from the datasets to be aligned. The features are geometrical entities, with the most commonly used ones being points, lines or contours~\cite{Saiti2020}.

\section{Particle Swarm Optimization}\label{pso}

The Particle Swarm Optimization (PSO) algorithm is a population-based optimization algorithm, proposed by Kennedy and Eberhart in 1995~\cite{Kennedy1995}. It is inspired by the behavior of swarms in nature. It works by iteratively improving a candidate solution by adjusting the values of the variables in the solution based on the experiences of the "particles" in the swarm.

The algorithm consists of a population of particles, each of which represents a potential solution to the optimization problem. Each particle has a position in the search space and a velocity, which determines how the particle moves through the search space. The particles are initialized at random positions in the search space, and the velocity of each particle is initialized to a small value.

At each iteration of the algorithm, each particle updates its position and velocity based on its current position, the best position it has found so far (referred to as the "personal best" position), and the best position found by any particle in the swarm (referred to as the "global best" position). The position and velocity of each particle are updated according to the following equations:
\begin{equation}
V(t+1) = w V(t) + c_1 r_1 (PBest(t) - P(t)) + c_2 r_2 (GBest(t) - P(t)) 
\end{equation}
\begin{equation}
P(t+1) = P(t) + V(t+1) 
\end{equation}


%
where:
$V$ is the velocity, $P$ is the position of the particle,
$t$ is the current iteration,
$w$ is the inertia weight, which determines how much the particle's current velocity affects its new velocity,
$c_1$ and $c_2$ are constants that control the influence of the personal best and global best positions on the particle's velocity,
$r_1$ and $r_2$ are random numbers between 0 and 1,
$PBest(t)$ is the best position the particle has found so far,
$GBest(t)$ is the best position found by any particle in the swarm.

The algorithm continues until a stopping criterion is met, such as a maximum number of iterations or a satisfactory solution being found.

PSO is a simple and effective optimization algorithm that has been applied to a wide range of problems in various fields. It is easy to implement and can find solutions quickly, making it a popular choice for many optimization problems.

Several variants of the original PSO has been proposed. In this review we will investigate if the standard PSO algorithm has been used or its variants.

See Nayak et al.~\cite{Nayak2022} for an in-depth analysis of PSO, its variants and its applications, and Gad~\cite{Gad2022} for a systemtic review on PSO, that also outlines previous review studies.

\section{Methodology and Results}\label{method}

We applied the systematic review methodology of Brereton et al.~\cite{Brereton2007}. In this section we describe how we planned, conducted and analyzed results. 

\subsection{Planning the review}\label{planning}

During this phase, we identified the needs for this literature review. We reviewed the state-of-the-art existing literature in this area. The research questions were generated accordingly, and they are listed as follows:

\begin{enumerate}
\item[RQ1:] What is the brief overview of current research?
\item[RQ2:] What are the extensively applied approaches?
\item[RQ3:] How the work is distributed according to time
division?
\item[RQ4:] What are the address registration problems?
\item[RQ5:] What are the applied algorithms?
\end{enumerate}

\subsection{Conducting the review}\label{conducting}

We defined the inclusion and exclusion criteria:
\begin{enumerate}
\item Papers must focus on PSO for 3D medical image registration.
\item Papers must be written in English.
\item Papers must describe the methodology. Review papers are excluded.
\item Only published or peer-reviewed works are included. Dissertations and theses are excluded.
\item Papers where PSO is not used for registration (i.e. for segmentation, feature extraction or classification)   are excluded.
\item Papers not on medical applications or not on human are excluded.
\item Low quality papers, i.e. missing parameters or not reproducible are excluded.
\end{enumerate}

We conducted the search in December 2022, using the following digital databases:
\begin{enumerate}
\item PubMed
\item IEEExplore
\item Elsevier Science Direct
\item Springer Link
\item Google Scholar
\end{enumerate}

The search string was {\it(("Particle Swarm Optimization" OR "PSO") AND "3D" AND "medical" AND "image registration")}.

After removing duplicated, we examined title, abstract and keywords to determine if the papers fulfill the inclusion and exclusion criteria.
Finally, we assessed and read the papers in full for eligibility and data extraction.

\section{Results}\label{resuls}

We obtained 2830 articles from Google Scholar, 261 from Elsevier Science Direct, 58 from IEEExplore, 228 from Springer Link, 21 from PubMed. After duplicates removal and screening based on title, abstract and keywords, 92 papers were assessed in full according the inclusion/exclusion criteria listed in Sec.~\ref{conducting}.

No time range was applied during the search in order to
include as many search results as possible. Although PSO
was introduced by Kennedy et al.~\cite{Kennedy1995} over 25 years ago, our review only managed to identify 24 relevant papers using PSO to optimize 3D medical image registration (see Table~\ref{Table:Papers}). Many papers were excluded because they described only 2D and not 3D registration.
From papers, we retrieved the following information:

\begin{enumerate}
\item Modalities used for the registration
\item Anatomical Region
\item Dataset used
\item PSO type (standard PSO, hybrid or variants)
\item Correspondence (intensity or feature based)
\item Similarity Metric
\end{enumerate}

We also extracted the performance of the various methods, but given the many different evaluation measures used, we could not compare the proposed methods.

\subsection{RQ1: Brief Overview of Ongoing Research}\label{RQ1}
The selected literature for this systematic review is summarised in Table~\ref{Table:Papers}. By analysing it we will overview the PSO techniques that are applied to register 3D medical images, as well as which are the common image modalities that are used in the selected papers.

\begin{table}[!h]
\caption{Selected papers}\label{Table:Papers}
\begin{tiny}
\begin{tabular*}{\textheight}{llllllll}
    \toprule
        {\bf Authors} & \multicolumn{2}{l}{\bf Modalities} & {\bf Anatom.} & {\bf Dataset} & {\bf PSO}  & {\bf Corresp.} & {\bf Similarity} \\ 
        & & & {\bf Region} & & {\bf Type} & & {\bf Metric} \\ 
        \midrule
        Talbi and Batouche & MRI & CT & brain & private & DEPSO & intensity & MI \\ 
        ~\hspace{1.8cm}~\cite{Talbi2004} & MRI & RX &  &  &  &  & \\
        & MRI & SPECT   &  &  & &  \\ 
        & PET & MRI &  &  &  & &  \\ 
        Wachowiak et al.~\cite{Wachowiak2004} & US & histology & abdomen & NLM-NIH & HPSO & intensity & NMI \\ 
        & CT & histology & brain & BrainWeb &  &  & \\ 
        & MRI\_T1 & MRI\_T2 & &  &  &  & \\ 
        Li et al.~\cite{Li2008}& MRI & CT & brain & private & PSO & points & LTS-HD \\
        Chen and Mimori~\cite{ChenMimori2009} & MRI   & CT & brain & RIRE & HPSO & intensity & MI \\ 
        Chen et al.~\cite{Chen2009} & MRI   & CT & brain & RIRE & HPSO & intensity  & MI \\ 
        Li et al.~\cite{Li2010} & CT & CT & brain & private & NCQPSO & intensity & JS \\ 
        & MRI & CT & &  &  &  & \\ 
        Zhang et al.~\cite{Zhang2010}& CT  & CT & brain & private & PSO & intensity & GMI \\ 
        & CT  & MRI &  &  &  & &  \\ 
        Bao and Sun~\cite{Bao2011} & CT  & MRI & brain & RIRE & RQPSO & intensity & MI \\ 
        & PET & MRI &  & &  & & \\ 
        Zhou et al.~\cite{Zhou2011} & CT  & MRI & brain & RIRE & RQPSO,  & intensity & NMI \\ 
        & PET & MRI &  &  & DRQPSO &  &  \\ 
        Ayatolahi et al.~\cite{Ayatollahi2012} & CT  & CT  & brain & AANLIB & HPSO & intensity & MNMI  \\ 
        &  MRI\_T2  &  MRI\_PD  &  &  &  &  &   \\ 
        &  MRI\_T2 &  CT &  &  &  &  &  \\ 
        Lin et al.~\cite{Lin2012}& MRI & CT & brain & RIRE & HPSO & intensity & MI \\ 
        Kang et al.~\cite{Kang2014} & 2D X-rays & 3D X-rays & femur & FTRAC & EM-PSO & geometry & MLE \\ 
        Schwab et al.~\cite{Schwab2015} & CT & MRI & brain & RIRE & 4 PSO var & intensity & NMI  \\ 
        Wang et al.~\cite{Wang2015}& MRI\_T2  & MRI\_T2 & cardiac & private & PSO & intensity & MI \\ 
        &  MRI\_T2 &  MRI\_DE &  &  &  &  &  \\ 
        &  MRI\_T1 &  MRI\_T2 &  neonatal &  &  &  &  \\ 
        Hui and Zhijun~\cite{Hui2015} & PET & MRI & brain & RIRE & QPSO & intensity & MI \\ 
        Manoj et al.~\cite{Manoj2016} & CT & MRI & brain & private & BAT-PSO & mixed & MI \\ 
        Hering et al.~\cite{Hering2016}& MRI & MRI & brain & private & MOPSO & intensity & TRE \\ 
        Hernandez-Matas et al. & fundus  & fundus  & retina & private & 8 PSO var & SIFT & ED \\ 
        \hspace{1.8cm}~\cite{Hernandez-Matas2017} & camera  & camera &  &  &  &  &  \\ 
        Li et al.~\cite{Li2017} & CT & US & femur & private & PSO & H-shaped im & MI \\ 
        Zaman and Ko~\cite{Zaman2018} & 3D CT & 2D X-rays & femur & private & PSO var & intensity & ED \\ 
        Wang et al.~\cite{Wang2019} & CT & MRI\_T2 & brain & RIRE & UKFPSO & intensity & TRE \\ 
        & PET & MRI\_PD  & brain  & RIRE &  &  &  \\ 
        & MRI\_T1 &  MRI\_T2  & neonatal & private &  &  &  \\ 
        Liu et al.~\cite{Liu2021} & CBCT & CBCT & dental & private & PSO & intensity & GMM-NCC \\ 
        Yoon et al.~\cite{Yoon2021} & CT & 2D X-rays & artery & private & MT-PSO & centerlines & CBKnn, CBO \\
        Shao et al.~\cite{Shao2022} & CT & CT & multiple  & private & PSO & points & ED \\ 
        \bottomrule
    \end{tabular*}
\hspace{0.5cm} \\
HPSO = Hybrid PSO \\
DEPSO = Differential Evolution PSO \\
QPSO = Quantum-Behaved PSO \\
NCQPSO = Niche Chaotic mutation Quantum-Behaved PSO \\ 
RQPSO = Revised Quantum-Behaved PSO \\
DRQPSO = Diversity Revised Quantum-Behaved PSO \\
MOPSO = Multi-Objective PSO \\
UKFPSO = Unscented Kalman Filter PSO \\
EM-PSO = Expectation Maximization PSO \\
MT-PSO = Multi-Threaded PSO \\
MI = Mutual Information \\
NMI = Normalized Mutual Information \\
MNMI = Modified Normalized MI \\
MLE = Maximum Likelihood Estimation \\
LST-HD = Least Trimmed Square Hausdorff Distance \\
JS = Jensen-Renyi measure \\
GMI = Generalized Mutual Information (based on Renyi Entropy) \\
TRE = Target Registration Error \\
ED = Euclidean Distance \\
CBKnn = Chamfer bijection distance with K-nearest neighbor pairing \\ 
CBO = Chamfer bijection distance with an orientation filter \\
GMM-NCC = Gaussian Mixture Model Normalized Cross-Correlation 
    \end{tiny}
\end{table}

\subsection{RQ2: Extensively Used Approaches}
\begin{figure}[h]
    \centering
    \includegraphics[trim={0 5cm 0 5cm},clip,width=0.7\textwidth]{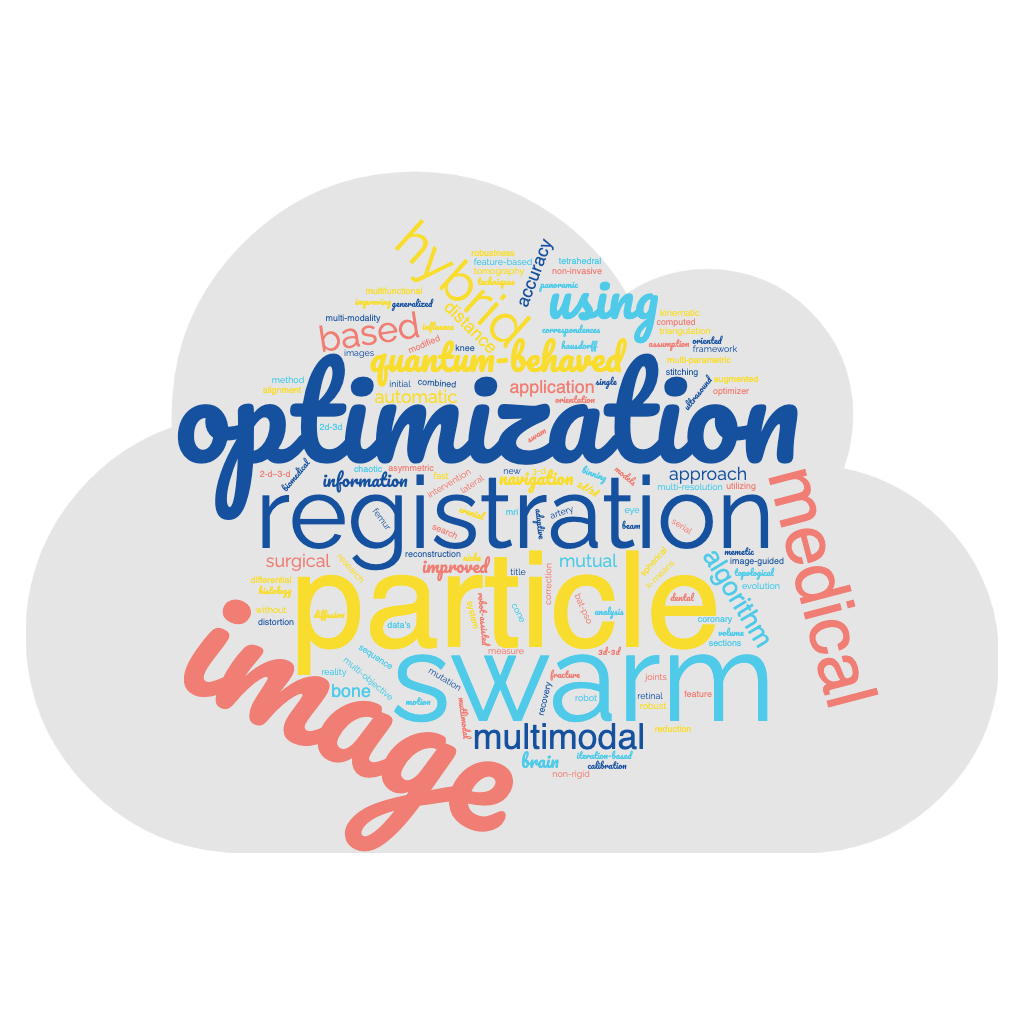}
    \caption{Wordcloud of Frequently Occurred Words in Titles of Selected Papers}
    \label{Fig:wordcloud}
\end{figure}

The wordcloud shown in Figure~\ref{Fig:wordcloud} is generated on the basis of terms enclosed in the titles of selected research articles. The size of each term demonstrates the number of their occurrence. 
According to the size of terms in Figure~\ref{Fig:wordcloud}, it can be observed that Particle Swarm Optimization, Medical Image Registration are the predominant words that frequently appear in the targeted research articles.

\subsection{RQ3: Literature Distribution with Time
Division}
\begin{figure}[h]
    \centering
    \includegraphics[width=0.7\textwidth]{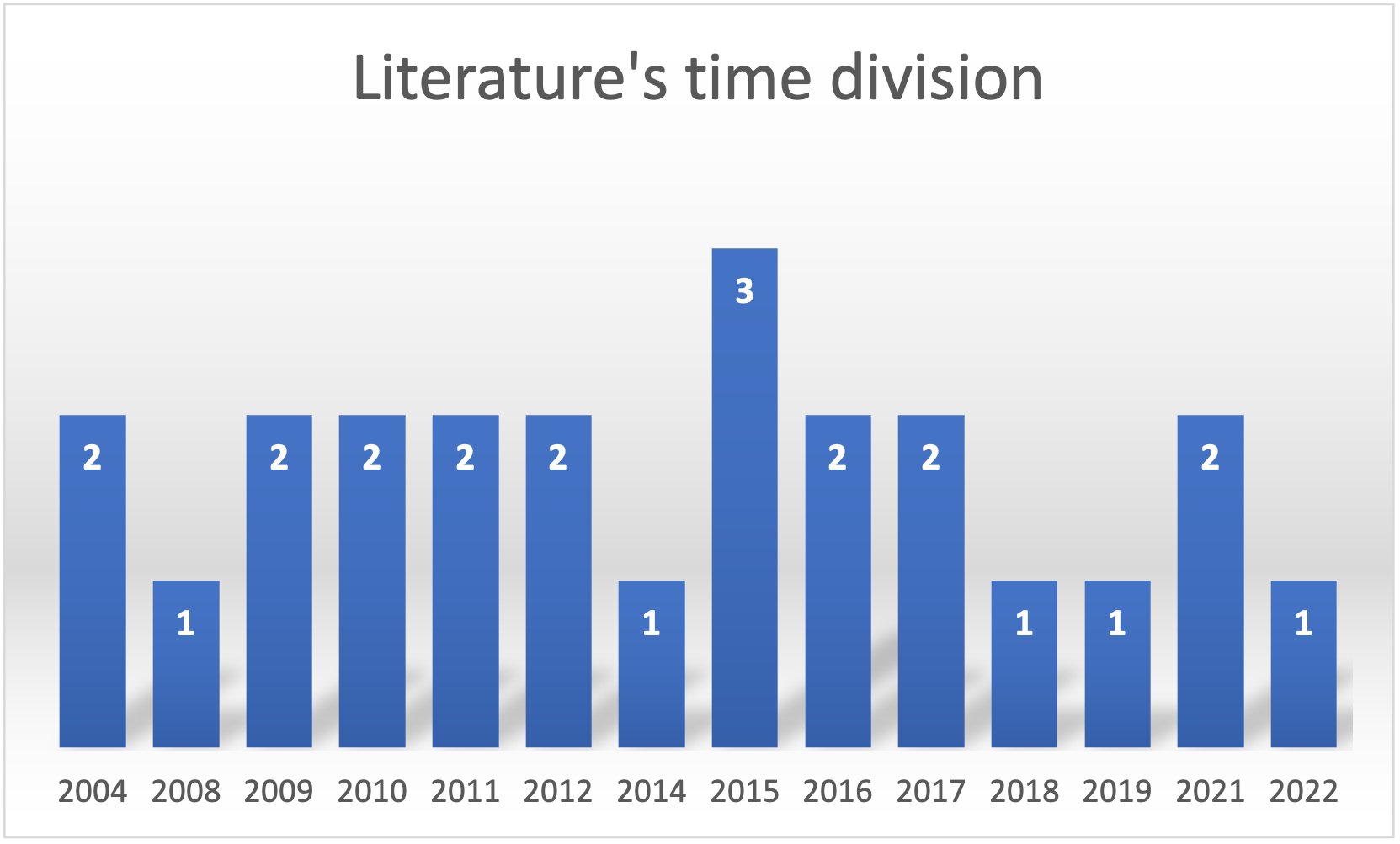}
    \caption{Temporal Distribution of Selected Papers}
    \label{Fig:time}
\end{figure}

The graph in Figure~\ref{Fig:time} illustrates the distribution of selected papers with respect to the time division. It shows that the number of papers published each year is quite constant. 

\subsection{RQ4: Addressed Registration Problems}

\begin{figure}[h]
    \centering
    \includegraphics[width=0.49\textwidth]{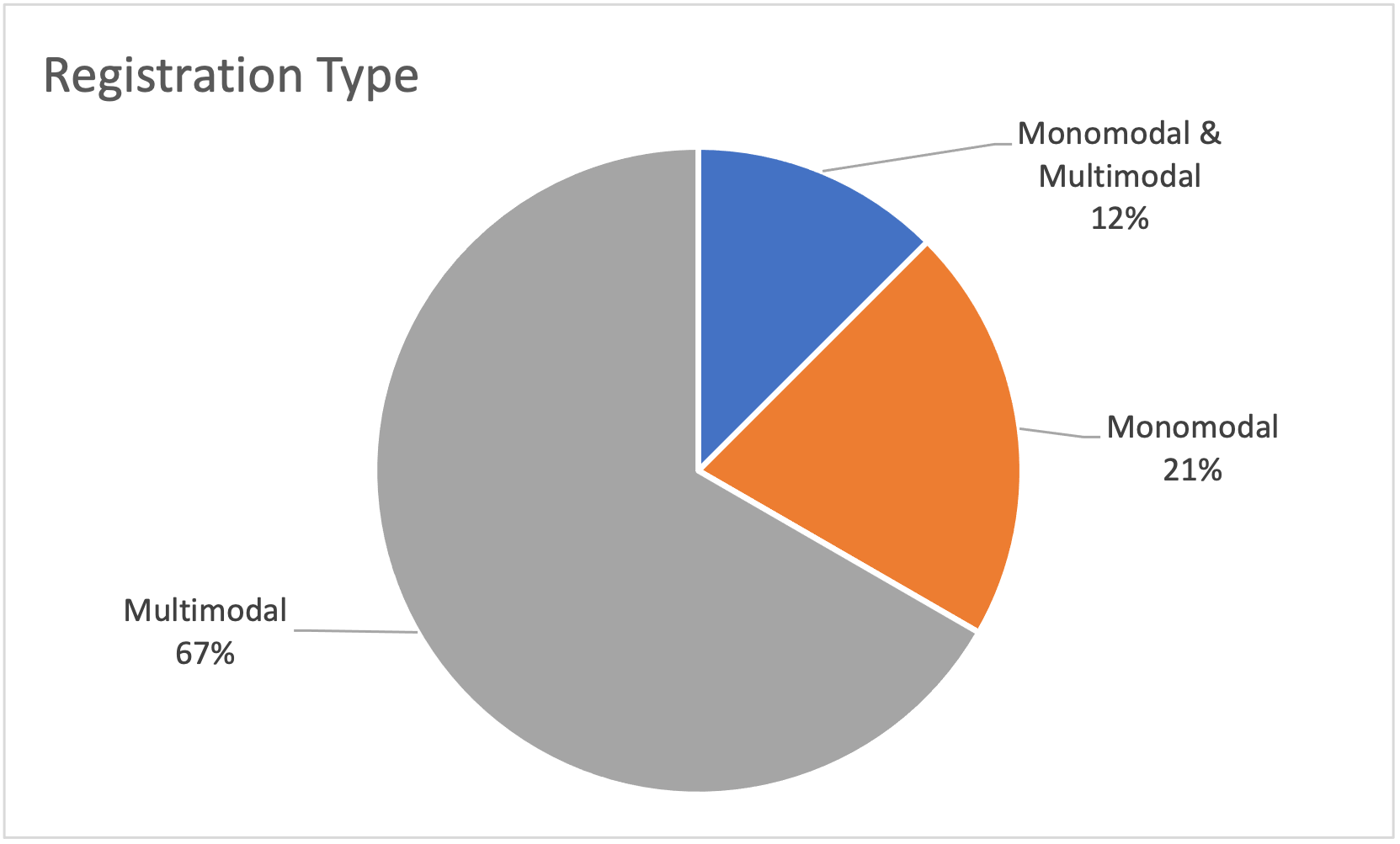}
    \includegraphics[width=0.49\textwidth]{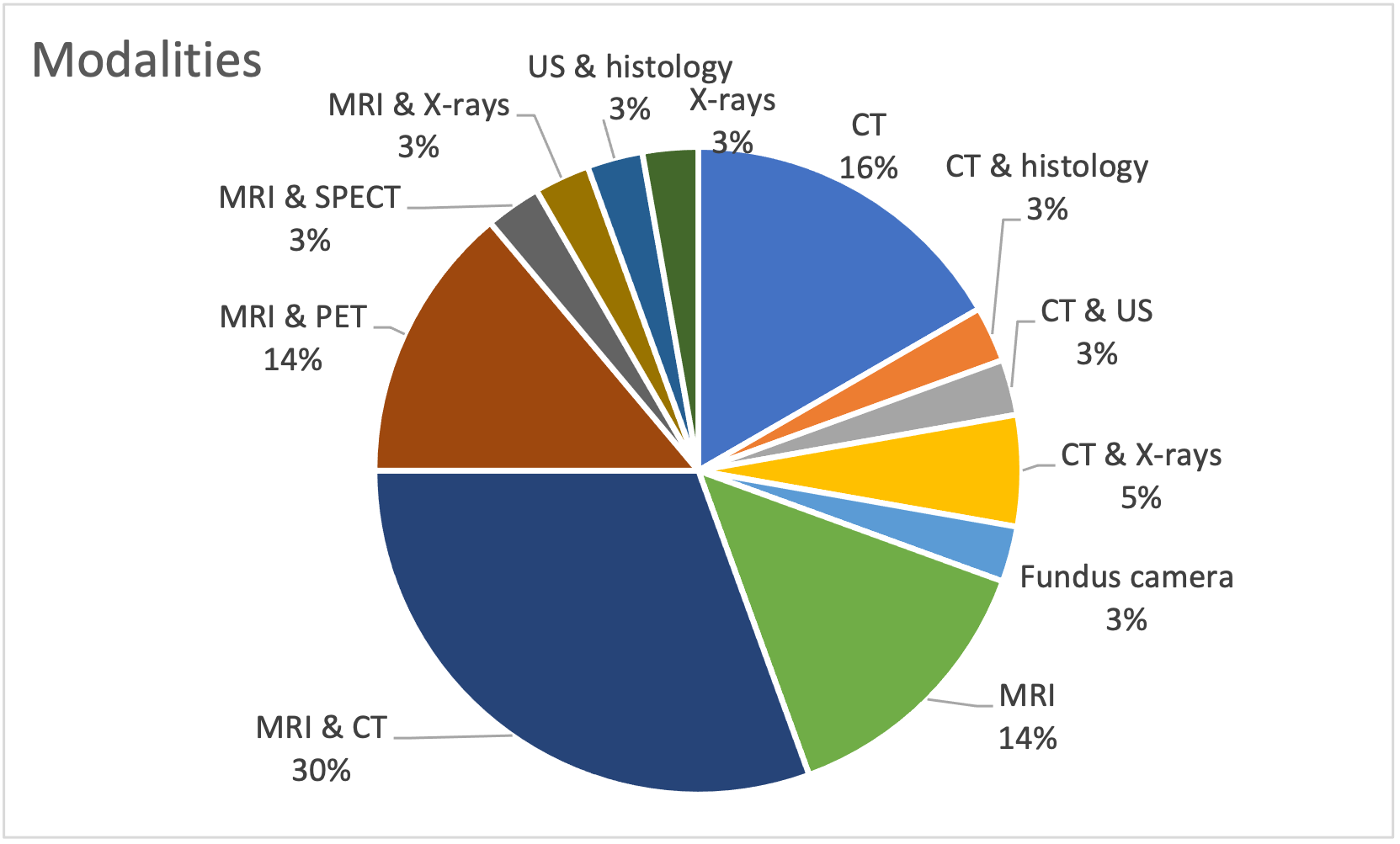}
    \caption{Registration type (monomodal or multimodal) and Registered Modalities}
    \label{Fig:modalities}
\end{figure}

Analysing Table~\ref{Table:Papers} some useful conclusions can be extracted: 67\% of the selected papers present examples of multimodal registration, 21\% of monomodal registration, and 12\% describe examples both monomodal and multimodal registration (see Figure~\ref{Fig:modalities}). In Figure~\ref{Fig:modalities} we can also observe which modalities are frequently used. The multimodal registration of Magnetic Resonance Image (MRI) and Computer Tomography (CT) is most commonly studied (30\%), followed by MRI and Positron Emission Tomography (PET) (14\%), and by different modalities of MRI (14\%), for instance T1-weighted and T2-weighted. For single mode registration CT is the most frequently used modalities (16\%). Only one or two studies used other modalities, like fundus camera, Ultrasounds (US), histology or X-rays.

\begin{figure}[h]
    \centering
    \includegraphics[width=0.49\textwidth]{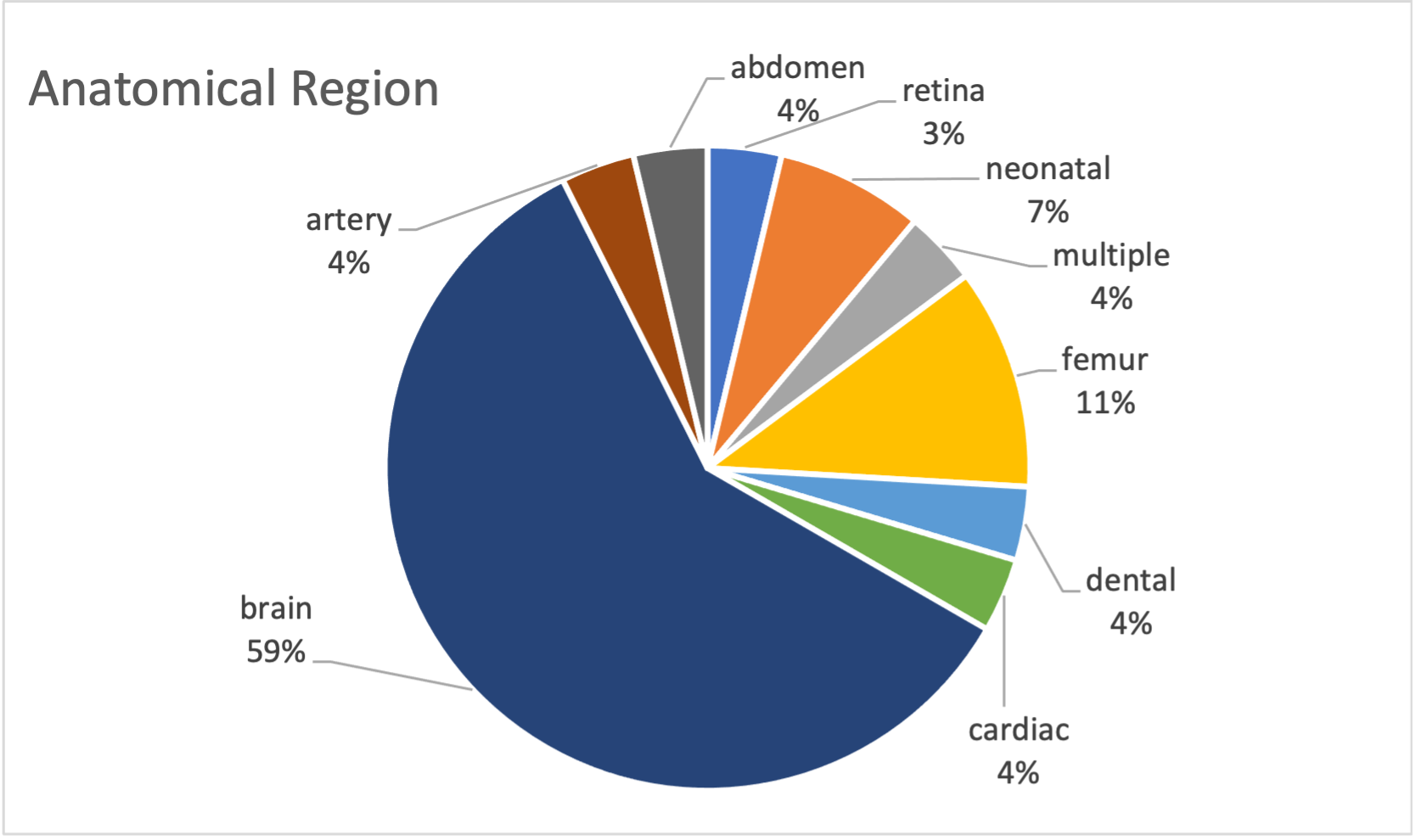}
    \includegraphics[width=0.49\textwidth]{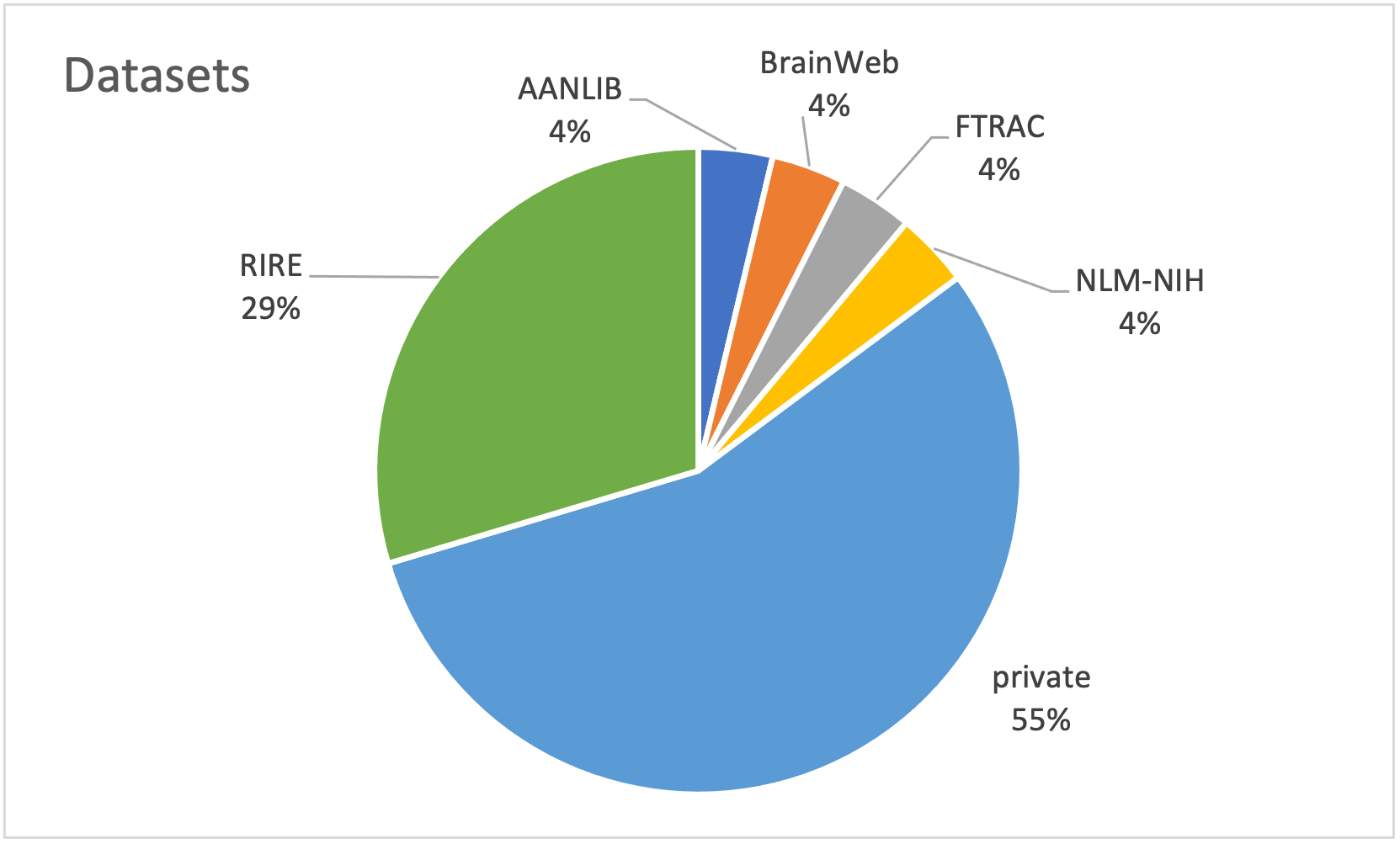}
    \caption{Anatomical Regions and Datasets used}
    \label{Fig:regions}
\end{figure}

Images of various anatomical regions have been studied, beeing the brain (59\%) the top one in terms of popularity (see Figure~\ref{Fig:regions}). 
Regarding the datasets upon which experiments were conducted by the presented techniques, it should be highlighted that 55\% are private while 45\% are publicly available (see Figure~\ref{Fig:regions}). The use of different datasets, makes comparison between the different approaches hard. The publicly available datasets are: RIRE~\cite{West1997,RIRE}, AANLIB~\cite{AANLIB}, BrainWeb~\cite{Kwan1999,BrainWeb}, FTRAC~\cite{FTRAC}, NLM-NIH~\cite{NLM-NIH}. 

\subsection{RQ5: Applied PSO Algoritms}

\begin{figure}[h]
    \centering
    \includegraphics[width=0.49\textwidth]{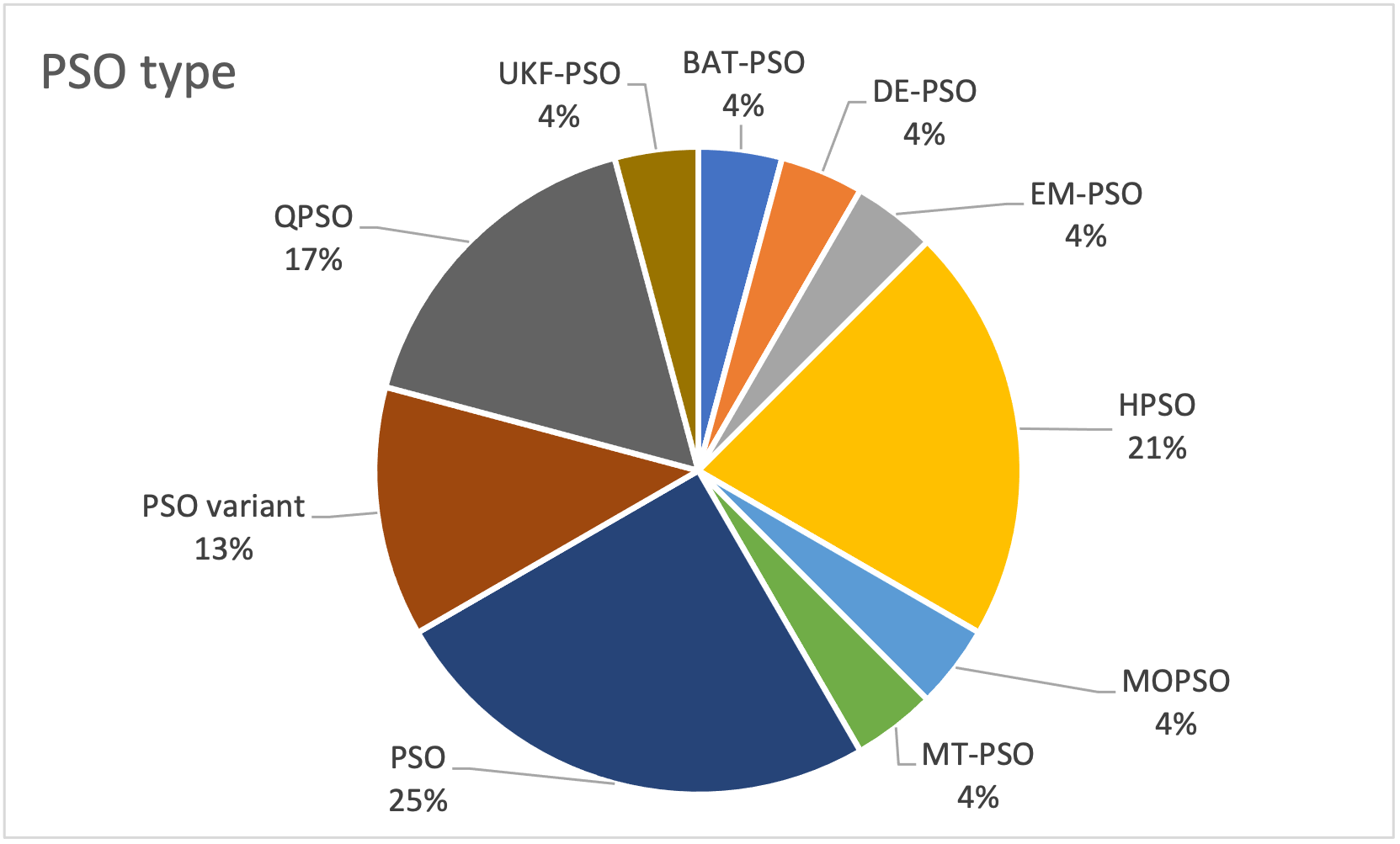}
    \includegraphics[width=0.49\textwidth]{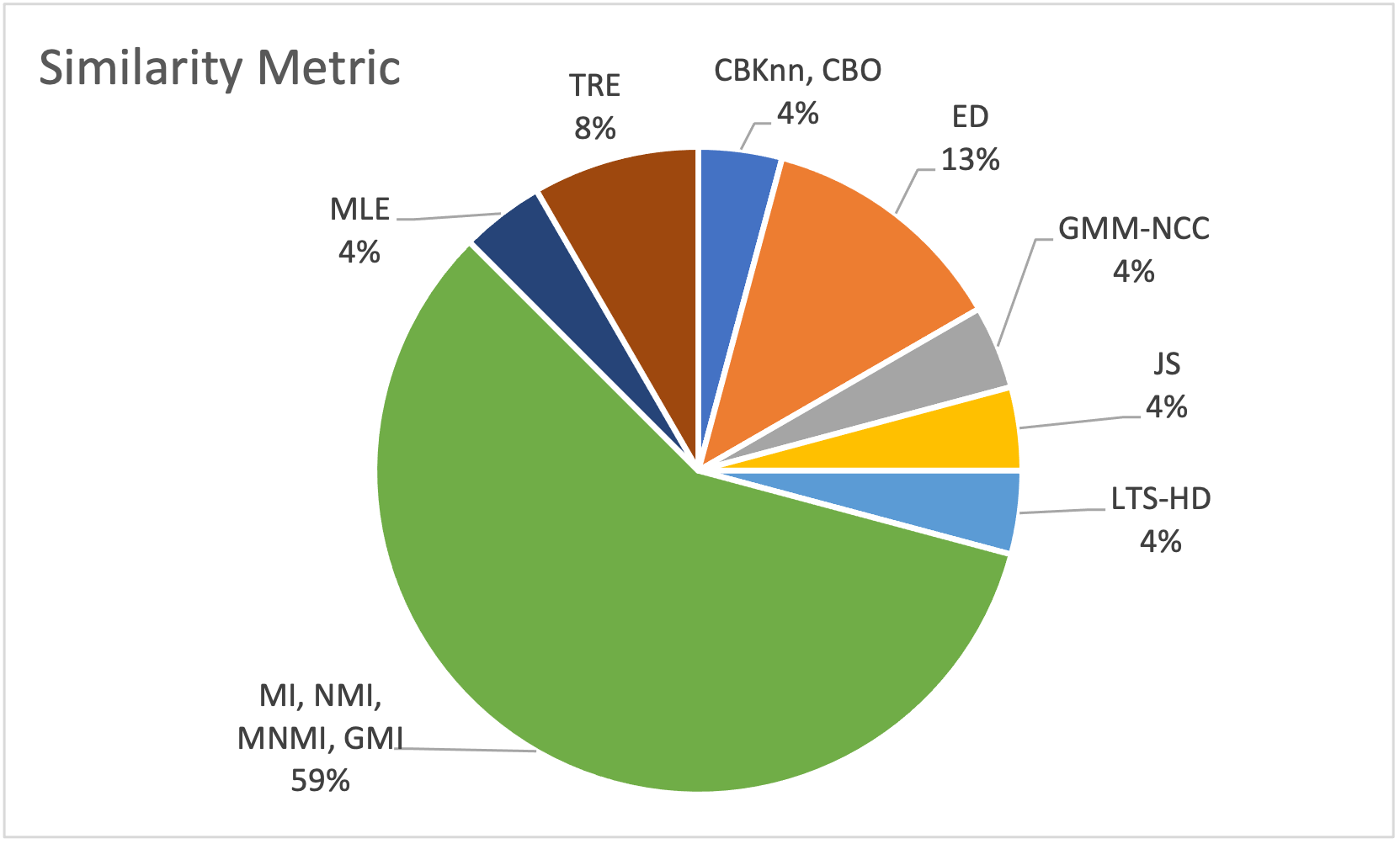}
    \caption{PSO Strategy (standard, improved variants or hybrid PSO) and Similarity Metric}
    \label{Fig:type}
\end{figure}

The applied PSO algorithms can be seen in Figure~\ref{Fig:type}, where we observe that 25\% of researchers applied Standard PSO and the remaining improved of hybrid version of PSO. Some researchers named their version, ad therefore we reported the given name in the Table and in the Figure, while others simply called it Hybrid PSO (HPSO) or PSO variant. The Quantum Behaved PSO (QPSO) and its modification (Niche Chaotic mutation QPSO (NCQPSO), Revised QPSO (RQPSO) and Diversity Revised QPSO (DRQPSO)) is the most popular one (17\%), while other hybrid versions of PSO often incorporate elements from Genetic Algorithms, like Crossover or subpopulations. 

According to the type of correspondence, among the selected papers 71\% of registration methods are intensity-based and 29\% are feature-based.

Various similarity measures have been used as objective function of the PSO algorithm (see Figure~\ref{Fig:type}), being Mutual Information (MI) including its variants (Normalized MI (NMI), Modified Normalized MI (MNMI) and Generalized MI (GMI)) the most popular (59\%); Euclidean Distance is mainly used for feature based registration. Target Registration Error (TRE)~\cite{Fitzpatrick2001} and other similarity metrics are also adopted.

\section{Conclusions}

We presented a systematic review of existing studies on the application of the standard PSO and its variants to the problem of 3D medical image registration. 
In order to perform the systematic review, the gaps in the literature are figured out and converted into five research 
questions. The findings of this systematic review depict that some researchers applied PSO and its hybrid versions to register anatomical regions using different modalities of medical images as MRI, CT and more.

\bibliography{exportPSOpapers,exportPSOreview}


\end{document}